\documentclass{article}

    \usepackage[final]{neurips_2020}

\title{Learning to Communicate with Strangers \\ via Channel Randomisation Methods}

\author{%
  Dylan Cope
  \\
  \texttt{dylan.cope@kcl.ac.uk} \\
   \And
   Nandi Schoots \\
   \texttt{nandischoots@gmail.com} \\
   \AND
   King's College London \& Imperial College London
}

\usepackage{enumitem}
\usepackage[english]{babel}
\usepackage{xparse}
\usepackage{url}

\usepackage{dsfont}
\usepackage{amsthm}
\newtheorem{definition}{Definition}

\usepackage{caption}
\usepackage{subcaption}
\usepackage{wrapfig}

\usepackage{amssymb}
\usepackage{amsmath}
\usepackage{mathtools}

\usepackage{algorithm}
\usepackage{algorithmicx}
\usepackage[noend]{algpseudocode}
\usepackage{changepage}
\usepackage{bbm}

\algrenewcommand\algorithmicindent{1.0em}

\usepackage[bottom]{footmisc}

\usepackage{todonotes}

\usepackage{etoolbox}
\newcommand{\numberthis}{\refstepcounter{equation}\tag{\arabic{equation}}}
\expandafter\appto\csname align*\endcsname{\numberthis}
\allowdisplaybreaks

\usepackage{booktabs}

\usepackage{dblfloatfix}

\usepackage[shortcuts]{extdash}
\DeclarePairedDelimiter\brac{(}{)}
\DeclarePairedDelimiter\curlybrac{\{}{\}}
\DeclarePairedDelimiter\sqbrac{[}{]}

\renewcommand{\vec}[1]{\ensuremath{\mathbf{#1}}}

\newcommand{\randvar}[1]{\ensuremath{\mathrm{#1}}}

\newcommand{\mathnull}{\ensuremath{\text{\O}}}

\DeclarePairedDelimiter\ceil{\lceil}{\rceil}

\usepackage{centernot}
\newcommand\independent{\protect\mathpalette{\protect\independenT}{\perp}}
\def\independenT#1#2{\mathrel{\rlap{$#1#2$}\mkern2mu{#1#2}}}
\newcommand\dependent{\ensuremath{\centernot\independent}}

\begin{document}




\maketitle 

\begin{abstract}
We introduce two methods for improving the performance of agents meeting for the first time to accomplish a communicative task. 
The methods are: (1) `message mutation' during the generation of the communication protocol; and (2) random permutations of the communication channel.
These proposals are tested using a simple two-player game involving a `teacher' who generates a communication protocol and sends a message, and a `student' who interprets the message.
After training multiple agents via self-play we analyse the performance of these agents when they are matched with a stranger, i.e. their zero-shot communication performance.
We find that both message mutation and channel permutation positively influence performance, and we discuss their effects.
\end{abstract}
\vspace{-2cm}



\vspace{1.5cm}
\section{Introduction}

Given an environment in which multiple learning agents are rewarded for completing tasks, communicative behaviour may emerge as a tool to achieve higher rewards. 
Broadly speaking, the study of such emergent communication addresses the question of which circumstances lead to communication as an instrumental strategy.
Another field of study is that of zero-shot learning, where systems are trained to perform well on a task that they have never seen before. 
At the conjunction of these fields there are two important forms of zero-shot performance in emergent communication: communicating previously unseen data \citep{DBLP:conf/iclr/ChoiLF18, DBLP:journals/corr/DauphinTHH14}; and communicating with previously unseen agents \citep{DBLP:journals/corr/abs-2003-02979}, i.e. strangers. 
In this work we will focus on the latter, and we will refer to this problem as zero-shot communication learning. 


We start with the observation that to communicate effectively with strangers one needs to establish a shared communication protocol (or enhance some shared base-protocol with context).
When two agents that have never met before create a novel communication protocol within an episode we call this \textit{intra-episodic communication protocol establishment}. Conversely, a protocol that is agreed upon between episodes is an \textit{inter-episodically established protocol}, or just a \textit{fixed protocol}. This distinction will be made more concrete in Section \ref{sec:protocol_establishment}. 
Our goal is to train agents that can intra-episodically establish a communication protocol with a stranger and use this protocol to cooperate on a shared task.

A survey of the relevant literature demonstrates that when two agents are trained together, the agents typically converge on an inter-episodically fixed communication protocol \citep{Foerster2016LearningLearning, DBLP:conf/aaai/MordatchA18, Lowe2019OnCommunication}.
In other words, the parameters of each agent's policy store the protocol itself, rather than the skill of generating and interpreting a protocol. 
Depending on the use case, this may be acceptable. However, in a zero-shot communication setting we should not expect effective cooperation unless the agents have happened upon the same fixed protocols by chance. When the space of possible protocols is large, this is unlikely.

This leads to the hypothesis that such fixed protocols can be prevented by randomising across the space of protocols that agents are exposed to during training. Our first proposal is \textit{message mutation} (Section \ref{sec:msg_mut_def}), where, alongside a specific set of training signals, agents should learn intra-episodic communication learning capabilities when their fixed protocol is randomly tampered with during an episode. Our second proposal is \textit{channel permutation} (Section \ref{sec:chan_perm_def}), where for each episode a random permutation map over the communication symbols is defined, and each time a symbol is sent through the communication channel it is transformed according to this mapping. 
Note that one instance of channel permutation (permutation with maximal subset size) is a form of \textit{Other Play} \citep{DBLP:journals/corr/abs-2003-02979}. 
We demonstrate that after training in either of these schemes the agents indeed have enhanced zero-shot communication performance (Section \ref{sec:experiments}).

In order to study this we introduce a simple environment in which agents only have communicative actions.
This allows us to isolate the effect of randomisation on ability to establish a communication protocol with a stranger.
In Section \ref{sec:discussion} we discuss how our proposals may generalise to more complex environments.
To our knowledge, our novel contributions are:
(1) introducing a formal distinction between intra-episodic and inter-episodic protocol establishment;
(2) the two proposed channel randomisation methods; and
(3) isolating the effects of these methods on zero-shot communication.

\section{Preliminaries}

\textbf{Protocols:}
The term ``communication protocol'' is used broadly and generally refers to ``any agreed upon set of behaviours that facilitates communication''. For our purposes, we define a communication protocol $p$ as a mapping from a set of \emph{subjects} $X$  to a set of communication \emph{symbols} $\Sigma$, i.e. $p : X \rightarrow \Sigma$. We will refer to elements of $\Sigma$ as \emph{utterances} or \emph{messages}, and $X$ will typically be a set of agent observations.  
We will denote messages from agent $j$ to agent $i$ at time $t$ with $m_{ijt}$. We denote observations of agent $i$ at time $t$ with $o_{it}$.
For brevity when we are agent-agnostic we will drop the $i$ and $j$ subscripts, i.e. we will use $m_t$ and $o_t$.

\textbf{Domain Randomisation:}
Domain randomisation is a training-time technique for improving zero-shot performance of a learning system when it is transported to a new domain \citep{8202133, DBLP:conf/corl/MehtaDGPP19, DBLP:journals/corr/abs-2003-02979}, which becomes relevant when the agent can not be directly trained in the target environment.
By randomising certain features of the training environment we apply pressure on a learning agent to find strategies that can adapt to changes in these features. 
As we are interested in zero-shot communication learning, our proposals involve introducing specific forms of randomisation into the communication channel that can achieve such results for the domain of possible communication protocols. 

\textbf{Markov Games:}
Markov games are an extension of (partially-observable) Markov decision processes to a multi-agent setting \citep{Littman1994MarkovLearning}. For $N$ agents, a Markov game is defined by the following components. Firstly, a set of environment states $\mathcal{S}$ and a probability distribution over initial states $\rho : \mathcal{S} \rightarrow [0, 1]$, For each agent $i$ there is: a set of actions $\mathcal{A}_i$, a set of observations $\mathcal{O}_i$, a function for extracting agent-dependent observations $\omega_i : \mathcal{S} \rightarrow \mathcal{O}_i$, and a reward function $r_i : \mathcal{S} \times \mathcal{A}_1 \times \dots \times \mathcal{A}_N \rightarrow \mathds{R}$. Finally, the environment dynamics are defined by a stochastic transition function $\mathcal{T} : \mathcal{S} \times \mathcal{A}_1 \times \dots \times \mathcal{A}_N \rightarrow \Delta(\mathcal{S}) $, where at time $t$ with state $s_t$ and agent actions $a_1, \dots, a_N$, the next state $s_{t+1}$ is sampled from the distribution: $s_{t+1} \sim \mathcal{T}(s_t, a_1, \dots, a_N)$. For finite Markov games we also select one or more states $\mathcal{S}_T \subseteq \mathcal{S}$ as terminal states. An `episode' of a Markov game is defined as the state and action history from the initial state to a terminal state. The function generating each agent's actions is called their policy, $\pi_i: \mathcal{O}_i \rightarrow \mathcal{A}_i$. In this work we consider cooperative games where each agent receives the same reward. 

In Markov games with communication, the action space $\mathcal{A}_i$ can often be expressed as a Cartesian product of environment actions $\mathcal{A}_i^e$ and communicative actions $\Sigma$, $\mathcal{A}_i = \mathcal{A}_i^e \times \Sigma$.

\section{Protocol Establishment} \label{sec:protocol_establishment}

We seek to analyse the performance on a communicative task of agents meeting for the first time. 
In order to communicate with a stranger, some form of protocol needs to be established. 
A protocol allows for a sender to signal information via a message, and for a receiver to listen to information and use it to inform its decisions.
We formalise these concepts in Section \ref{sec:eme-com}.
We consider agents to be strangers when they meet for the first time in an episode.
In Section \ref{sec:intra-episodic} we introduce intra-episodic protocol establishment.
All these concepts are relatively general. 
In Section \ref{sec:measures} we will introduce the more specific concepts that we use in our setting.







\subsection{Measuring Emergent Communication}\label{sec:eme-com}


\cite{Lowe2019OnCommunication} discuss the pitfalls of measuring the presence of emergent communication and identify two important behaviours: \textit{positive signalling} (see Definition \ref{def:positive_signalling}) and \textit{positive listening} (see Definition \ref{def:positive_listening}). Positive signalling is when a message is correlated with some observation or intended action, and positive listening is when an agent changes their beliefs or behaviour in response to receiving such signals. The authors show that there are circumstances under which positive signalling is learned, but positive listening is not. We build on these definitions and add the definitions of \emph{positive signalling a subset}, \emph{simplified positive listening}, and \emph{positive listening with sensitivity}.

\begin{definition}[Positive Signalling \citep{Lowe2019OnCommunication}] \label{def:positive_signalling}
    Given a sequence of messages $\mathbf{m}$ sent, observations $\mathbf{o}$ made, and actions $\mathbf{a}$ taken by an agent over the course of a trajectory of length $T$, the agent is \textbf{positive signalling} if $\vec{m}$ is statistically dependent on $\vec{a}$ and/or $\vec{o}$.
\end{definition}

%
For our purposes, we are not interested in positive signalling only,
we are interesting in capturing which prior observation-message pairs are influencing current behaviour.

\begin{definition}[Positive Signalling a Subset] \label{def:positive_subset_signalling}
    Let $V_t$ be the set of observations made and actions taken up until, and including, timestep $t$.
    We say a message $m_{t+1}$ is \textbf{positive signalling a subset} $W \subset V_{t}$ if the message is dependent on $W$ even when we already condition on its complement, i.e. the following holds:
    \[
        P(m_{t+1} ~|~V_{t}) \neq P(m_{t+1} ~|~ V_{t} \setminus W).
    \]
    
\end{definition}

A specific case of positive signalling a subset is \textbf{positive signalling a pair} $(o_t = o, m_t = m)$ in message $m_{t+k}$. We will use this concept later to help define signalling a protocol within an episode.


\begin{definition}[Positive Listening \citep{Lowe2019OnCommunication}] \label{def:positive_listening}
    Given a set of messages $\Sigma$, an agent $i$, following policy $\pi: \mathcal{O}_i \times \Sigma \rightarrow \mathcal{A}_i$, is \textbf{positive listening} to another agent $j$ at time $t$ if $j$ has just sent message $m_{ijt} \in \Sigma$ to $i$ and $\|\pi(o_{it}, m_{ijt}) - \pi(o_{it}, m_{\mathnull})\|_\tau > 0$,
    where $\| \cdot \|_\tau$ 
    is a distance in the space of expected trajectories followed by $\pi$, 
    and $m_\mathnull \in \Sigma$ is a special silence message, e.g. a zero-vector. 
\end{definition}


Assuming the Markov property, 
the probability distributions over trajectories starting in state $s_t$ with respective actions $a_1$ and $a_2$ can only be different if $a_1 \neq a_2$. 
Note that the messages that agent $j$ received only affect the environment dynamics insofar as the agent's policy is affected.
Hence, 
the two probability distributions over trajectories following policy $\pi$ 
starting in state $s_t$
can only be different if $\pi(o_{it}, m_{ijt}) \neq \pi(o_{it}, m_{\mathnull})$.

\begin{definition}[Simplified Positive Listening] \label{def: simplified positive_listening}
    Given a set of messages $\Sigma$, an agent $i$, following policy $\pi: \mathcal{O}_i \times \Sigma \rightarrow \mathcal{A}_i$, is \textbf{positive listening} to another agent $j$ at time $t$ if $j$ has just sent message $m_{ijt} \in \Sigma$ to $i$ and $\pi(o_{it}, m_{ijt}) \neq \pi(o_{it}, m_{\mathnull})$.
\end{definition}

Note that in implementations in which some (communicative) actions are not directly accessible, 
the simplified positive listening criterion may actually be harder to evaluate than the original positive listening criterion.
Positive listening is a relatively weak demand. 
We can increase the demand by increasing the number of inequalities.

\begin{definition}[Positive Listening with Sensitivity] \label{def: sensitive listening}
    Given a set of messages $\Sigma$, an agent $i$, following policy $\pi: \mathcal{O}_i \times \Sigma \rightarrow \mathcal{A}_i$, is \textbf{positive listening with sensitivity} $1 \leq s \leq |\Sigma|$ to another agent $j$ at time $t$ if $j$ has just sent message $m_{ijt} \in \Sigma$ to $i$ and $\pi(o_{it}, m_{ijt}) \neq \pi(o_{it}, m')$ for at least $s$ different $m' \in \Sigma$.
\end{definition}

When the sensitivity is $\min \{|O_i|, |\Sigma|\}$, we say that the agent is \textit{sensitively listening}.
When a sender can positively signal many different pairs, and when a receiver can listen with high sensitivity, we can say that a \textit{rich protocol} 
is being used.

\subsection{Intra-Episodic and Inter-Episodic} 
\label{sec:intra-episodic}

We are interested in distinguishing between intra- and inter\-/episodically established protocols. 
Without any kind of `memory' there is no mechanism by which the agents can learn a protocol within an episode. 
We allow for memory by giving the agent a `mental state', which is blank at the start of each episode and which allows the agent to keep track of information across timesteps. In our experiments, this is implemented as the internal state of an LSTM Network (see Section \ref{sec:agent_architecture}). 
The agent can take the action of `sending' a mental state to itself, which becomes part of the agent's observation in the next state.
For each agent $i$, let there be a set $\mathcal{Z}_i$ of mental states, such that the agent's policy now has the signature: $\pi_i : \mathcal{O}_i \times \mathcal{Z}_i \rightarrow \mathcal{A}_i \times \mathcal{Z}_i$. 

We characterise intra-episodic protocol establishment as a process whereby a listener's `attention is drawn' to the subject of the protocol because of a recently received message \textit{and} an earlier exchange within an episode. In other words, after the protocol establishment, for a protocol $p$ mapping message $m$ to observation $o$ (where observations are the subject of $p$), receiving $m$ draws the agent's attention to $o$.
We formalise the concept of the agent's `attention being drawn' via the notion of mutual information \citep{DBLP:journals/bstj/Shannon48} between the agent's mental state and an earlier observation. 



\begin{definition}[Rise in mutual information]
Let $o_{l}$ be the observation of an agent in timestep $l$. Let $z_{t}$ be this agent's mental state at timestep $t$. 
Denote the mutual information between this agent's mental state at timestep $t$ and the observation at timestep $l$ with $I(z_{t};o_l)$. 
Then we define the rise of mutual information between these mental states and the observation as 
$\randvar{I}_{z_{t}o_l} :=I(z_{t+1};o_l) - I(z_{t};o_l)$. 
\end{definition}

With this we define intra-episodic protocol establishment formally as follows.

\begin{definition}[Intra-episodic protocol establishment] 
\label{def:intra_episodic}
    If an agent's rise in mutual information $\randvar{I}_{z_{t+k}o_t}$ is positive and causally influenced by $m_t$ and $m_{t+k}$,
    conditional on $m_{t+k} = m_t$, 
    i.e. the following holds
    \begin{equation}
    \begin{split}
         \randvar{I}_{z_{t+k}o_t} &> 0; ~\text{and} \\
        \randvar{I}_{z_{t+k}o_t} &\dependent m_{t+k}, m_t ~|~ m_{t+k} = m_t,
    \end{split}
    \end{equation}
    then we say that a protocol $p_{ij}(o) = m$ is established intra-episodically for message $m \in \Sigma$ and observation $o \in X$, 
    where $m = m_{t}$ and $o = o_t$.
\end{definition}

In the above definition, if agent $i$'s message $m$  
is statistically dependent on the observation action tuple $(o_{it}, m_{it})$, 
then we say that $i$ is positively signalling this pair (Definition \ref{def:positive_subset_signalling}). 
We say that agent $j$ is positively listening to observation message pair $(o_{jt}, m_{it})$, if its policy is affected by this pair (Definition \ref{def: simplified positive_listening}). 

While positive listening is an ingredient in intra-episodic protocol establishment, it is possible for there to be the former without the latter.
If an intra-episodic protocol has been established,
then the internal state $z_{t+k}$ is affected by message $m_{t+k}$, which means that the agent is positively listening to $m_{t+k}$.
However, this may be the case without any protocol having been established within this episode if the agent's policy is affected by a message but is unaccompanied by any precedent at an earlier timestep. 
We define inter-episodic protocol establishment as positive listening in the absence of intra-episodic protocol establishment.
We also refer to this phenomenon as \textit{protocol fixing}.
\begin{definition}[Inter-episodic protocol establishment]
\label{def:inter_episodic}
    A protocol $p_{ij}(o) = m$ is established inter-episodically for message $m$ and observation $o$ between agents $i$ and $j$ if 
    there is a timestep $t$ at which 
    $j$ is positively listening to observation message pair $(o_{jt}, m_{it})$, where $o_{jt}=o$ and $m_{it} = m$,
    and there is no intra-episodically established protocol $p_{ij}(o) = m$. 
\end{definition}

\section{Experimental Setting}

\subsection{Environment}

In order to design a minimal environment to test the emergence of intra-episodic communication learning we identify two key phases that should be present within each episode: a \textit{protocol establishment phase}, and a \textit{utilisation phase}. 
During the first phase the agents must have the opportunity to work together to associate observations and symbols. In the second phase they use this protocol to communicate information. 
For more complex environments agents could iteratively move between these two phases, but we consider a simple situation with one cycle of this process. 
In our environment there are two roles that an agent can play, which we refer to as the `teacher' and `student' roles. As both agents have the same architecture (described in Section \ref{sec:agent_architecture}), and as we will end up training agents by self-play \citep{Tesauro1994TD-GammonPlay} in this environment we will assume that any agent can play either of these roles.

As we are uninterested in exploring the capacity for learning agents to extract useful features from high-dimensional, complex inputs (such as images), we opt for a simple set of possible observations to be the subject of our agents' communication protocols\footnote{See the Supplementary Material for further justification of these design decisions.}. 
We call these the \textit{environment observations}, $\mathcal{O}_E$. 
The set $\mathcal{O}_E$ consists of numbers expressed as binary vectors.
Specifically, given $M$ desired classes, the environment observations are constructed as follows: 
\begin{align} \label{eqn:environment_obs_def}
    \mathcal{O}_E = \curlybrac*{(x_1, \dots, x_k) \in \mathds{Z}_2^k ~:~ 0 < y \leq M,~ {\textstyle \sum_i^k} 2^ix_i = y}.
\end{align}
Where $k = \ceil*{\log_2 M}$.
Additionally, we assign classifications to each member of $\mathcal{O}_E$ according to the number represented in binary (i.e. $y$ in Equation \ref{eqn:environment_obs_def}).

\begin{figure}[!t]
    \centering
    \scalebox{0.6}{\input{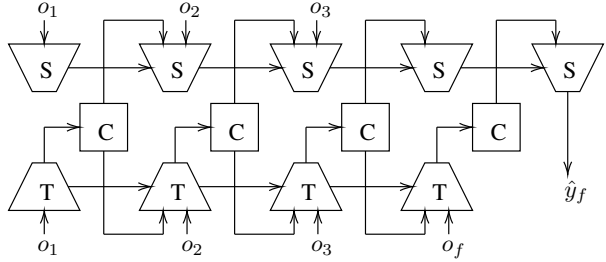}}
    \caption{
        Diagram of the information flow through an episode where $|\mathcal{O}_E| = 3$. S represents the student's policy, T represents the teacher's policy, and C represents the communication channel. Each $o_t$ is the input observation at time $t$, and $o_f$ is the final input that is hidden from the student. The output $\hat{y}$ at the final timestep is the student's prediction of $o_f$. The lateral connections between S-boxes and T-boxes show the information flow through the agents' mental states.
    }
    \label{fig:episode_arch}
\end{figure}

Figure \ref{fig:episode_arch} shows the information flow through an episode.
Each episode of our environment consists of $|\mathcal{O}_E| + 2$ timesteps. In the first $0 \leq t < |\mathcal{O}_E|$ timesteps, both the teacher and the student make the same environment observations $o_t$, and the teacher sends symbol $s_t \in \Sigma$ to the student, i.e. the teacher is given the opportunity to teach the student a communication protocol $p: \mathcal{O}_E \rightarrow \Sigma$. In the second phase, the \textit{testing phase}/utilisation phase, the teacher makes an observation that is hidden from the student and the teacher must communicate this information to the student. This phase consists of two time steps: one for the teacher to make an observation and produce a message, and another for the student to receive the message and produce a prediction. The agents' performance is measured as the mean classification accuracy of the student's predictions of the teacher's observation in the final timestep\footnote{See the Supplementary Material for a visualisation of good and bad performance in this environment, and for a formal construction of the environment.}. 


The game that we use can be interpreted as a form of the Lewis signalling game \citep{Lewis-convention, DBLP:conf/nips/ChaabouniKDB19, DBLP:conf/iclr/Lowe0FKP20}.
Alternatively it can be viewed as a referential game where the set of distractors coincides with the total set. Many works on observing emergent communication in referential games exist \citep{DBLP:conf/iclr/LazaridouHTC18, DBLP:conf/nips/HavrylovT17, DBLP:conf/iclr/EvtimovaDKC18, DBLP:journals/corr/abs-2001-03361, DBLP:journals/isj/CharafRH13}.

\subsection{Communication Channel} \label{sec:comm_channel}

During each episode our agents send messages to one another through a communication channel. At each timestep agents produce \emph{utterances} as log-probabilities (logits) over a discrete set of symbols $\Sigma$. These utterances are sent through the communication channel to produce \emph{messages}. We adopt a \emph{Differential Inter-Agent Learning} (DIAL) framework \citep{Foerster2016LearningLearning} for our experiments, meaning that the communication channel must be differentiable during the training phase. This is achieved by using the utterance logits to create a Gumbell-Softmax distribution, and sampling messages from this distribution \citep{Jang2017CategoricalGumbel-Softmax, Maddison2016TheVariables}. This requires setting a temperature parameter which we fix at 1.0 for all of our experiments (unless otherwise stated). In order to further encourage discretisation of the utterances we also follow the lead of \cite{Foerster2016LearningLearning} and inject noise into the channel (before sampling messages); we apply additive white Gaussian noise with a fixed standard deviation of 0.5. During the test evaluations, messages are constructed by computing the one-hot encoding of the argmax of the utterance logits.

\subsection{Agent Architectures} \label{sec:agent_architecture}

Each agent's policy network takes three inputs at each timestep and produces two outputs. The inputs are: (1) a one-hot encoding of the agent's most recently sent message, (2) a one-hot encoding of the other agent's most recently sent message, and (3) an environment observation $o \in \mathcal{O}_E$. The outputs are: (1) an utterance to send as a message through a communication channel, and (2) a probability distribution over the classes of possible observations. The agents' produce these actions according to their policy: an LSTM \citep{Hochreiter1997LongMemory} recurrent neural network parameterised by $\theta_i$, $\pi_{\theta_i} : \mathds{R}^{|\Sigma|} \times \mathds{R}^{|\Sigma|} \times \mathcal{O}_E \rightarrow \mathds{R}^{|\Sigma|} \times \mathds{R}^{|\mathcal{O}_E|}$.

\subsection{Message Mutation} \label{sec:msg_mut_def}

The first of our proposals for improving zero-shot communication is \emph{message mutation}. This is a function $f_m : \Sigma \rightarrow \Sigma$ defined as follows: 
\begin{equation}
\begin{gathered}
    f_m(s) = \begin{cases}
        s' & \text{if}~x < p_m \\
        s & \text{otherwise}
    \end{cases} \\
    \text{where}~x\sim \text{Uniform}([0, 1]) ~\text{and}~ s' \sim \text{Uniform}(S).
\end{gathered}
\end{equation}
Where $p_m$ is the \textit{mutation probability} and $S \subseteq \Sigma$ is the set of possible symbols that can be chosen from. How $S$ is constructed leads to the distinction between \emph{kind} and \emph{unkind} mutations. For kind mutations, $S = \{s \in \Sigma ~:~ s \notin H\}$, where $H$ is the history of sent messages, and for unkind mutations $S=\Sigma$. The motivation for this distinction is to avoid mutations unfairly impacting the teacher's attempt to produce consistent communication protocols when $\Sigma$ is small.

\subsection{Channel Permutation} \label{sec:chan_perm_def}

The second of our proposals is \textit{channel permutation}. Permutation is a Boolean variable: if it is enabled then for each episode an arbitrary bijective total function $f_{ij}: \Sigma \rightarrow \Sigma$, is created for every possible ordered pair of agents $i$ and $j$. More precisely, we sample from a uniform distribution over the symmetric group $S_{|\Sigma|}$:
\begin{align}
    f_{ij} \sim \text{Uniform}(S_{|\Sigma|}).   
\end{align}
Consequently, whenever agent $i$ sends a symbol $s \in \Sigma$ to agent $j$, agent $j$ receives $f_{ij}(s)$.

We also investigate only permuting a subset of the symbols, i.e. for a subset size $k$, we uniformly sample without replacement a subset $S \subseteq \Sigma$, and instead sample the map as follows:
\begin{align}
    f_{ij} \sim \text{Uniform}\brac*{\curlybrac*{
        f \in S_{|\Sigma|} ~:~ f(x) = x ~\forall x \notin S
    }}.
\end{align}

We refer to this variant as \textit{channel subset permutation}.

\subsection{Difference between Message Mutation and Channel Permutation}

We expect that channel permutation and message mutation will have somewhat different effects on the teacher, but roughly the same effect on the student.
In both approaches the teacher makes utterances which may or may not be changed (mutated or permuted) to a different message. However, in message mutation the teacher must react to changes in their protocol and adapt their behaviour in the final timestep when it needs to use the protocol. If the teacher wants to send the same message that the student received, then it needs to keep track of how the protocol was changed.

On the other hand, in channel permutation every utterance is consistently permuted, so if in the final timestep a teacher wants to communicate the same message as before, then they just have to make the same utterance as before.
The teacher does not have to be adaptive and may converge on a specific protocol. The student on the other hand is met with many different protocols.
From this perspective, it appears that message mutation is strictly better than channel permutation, in the sense that it encourages more adaptive behaviour from both agents.
However, message mutation comes with the requirement that the agents are (indirectly) rewarded for paying attention to the protocols established within the episode.
Permutation does not have this requirement, and can be implemented without explicit reference to a protocol. 


\subsection{Measures} \label{sec:measures}

Suppose that at time $|\mathcal{O}_E|$ (the testing phase) the teacher made observation $o_f$ of class $\vec{y}_f$, where $\vec{y}_f$ is a one-hot encoding of a class label, and made utterance $\vec{u}_f$. At $t<|\mathcal{O}_E|$ (the protocol establishment phase) they made observations $o_t$ of class $\vec{y}_t$ and made utterances $\vec{u}_t$. Suppose further that the student receives messages $\vec{m}_t$ at each timestep $t$ and outputs  $\vec{\hat{y}}_f$ in the final timestep.

\subsection*{Error Metrics}

\textbf{Actual-Class (AC) error:} the categorical cross-entropy ($CCE$) between the student's predictions and the actual class of $o_f$:
\begin{align}
    \mathcal{L}_{AC} = CCE(\vec{\hat{y}}_f, \vec{y}_f).
\end{align}

\textbf{Student-Implied-Class (SIC) error:} the categorical cross-entropy between the student's predictions and the predictions that they ought to have made given the protocol established in the episode:
\begin{align}
\begin{gathered}
    \mathcal{L}_{SIC} = CCE(\vec{\hat{y}}_f, \vec{y}^*), \\
    \text{where}~ \vec{y}^* = \begin{cases}
        \frac{1}{|T|} \sum_{t \in T} \vec{y}_t & \text{if}~|T|\geq 1 \\
        \text{Uniform}(\mathcal{O}_E) & \text{otherwise}
    \end{cases} \\
    \text{and}~ T = \curlybrac*{t < |\mathcal{O}_E| ~:~ \vec{m}_t = \vec{m}_f} .
\end{gathered}
\end{align}

\textbf{Teacher-Message (TM) error:} the categorical cross-entropy between the utterance that the teacher made and the message they should have sent:
\begin{align}
    \mathcal{L}_{TM} = CCE(\vec{u}_f, \vec{m}_t),~ \text{where}~ o_t = o_f.
\end{align}

A low TM error corresponds to positive signalling the pair $(m_t, o_t)$.


\textbf{Protocol Diversity (PD) error:} Given a matrix $P$ with $|\mathcal{O}_E|$ rows and $|\Sigma|$ columns, where each row $i$ corresponds with the message sent by the teacher at $t=i$, we define the protocol diversity error as:
\begin{align}
    \mathcal{L}_{PD} = \max_{\vec{c}_i} \|\vec{c}_i\|_1
    ~~~~~ \text{where}~ P = \sqbrac*{\vec{c}_1, \dots, \vec{c}_{|\Sigma|}}.
\end{align}
By virtue of the communication channel the rows of $P$ are normalised, meaning that $\mathcal{L}_{PD}$ is bounded by 1 and $|\mathcal{O}_E|$. 
When this metric is 1, the protocol being expressed is injective. The higher the error gets the more ambiguous the protocol is, keeping in mind that each row of the $P$ matrix is ideally a one-hot vector, but during training the distribution could be imperfect (see Section \ref{sec:comm_channel}).

\subsection*{Evaluation Metrics}

\textbf{Performance:} We measure (cooperative) performance of agents as the classification accuracy of the student's prediction with respect to $o_f$. In other words, the number of times that the student made the correct prediction divided by the number of predictions that it was asked to make. 

If the performance is high the teacher is \textbf{positively signalling} $o_f$ through $m_f$ and the student is \textbf{positively listening} to $m_f$. This must be the case due to the fact that the only information channel between $o_f$ and the student's prediction is $m_f$.

To use these metrics to detect intra- vs. inter-episodic protocol establishment, we can start by noting that high performance in the presence of a high TM error, high SIC error, or high PD error implies that the agents have converged on a fixed protocol, 
i.e. that they are using an \textbf{inter-episodically established protocol}. 
For example, given a fixed injective protocol $p_{ij}$, assume that the teacher sent the same message $m$, such that $m \notin \text{Image}(p_{ij})$, every timestep until the final timestep. For the final time step they make observation $o_f$ and send message $p_{ij}(o_f)$, which results in high TM error. The student then correctly interprets this message, but receives high SIC error, and the agents achieve high performance. Additionally, because all of the messages sent during the protocol establishment phase are the same, the PD error is also maximised. 

However, it is possible for all these error metrics to be low, and the performance metric to be high, and yet there still to be a fixed protocol $p_{ij}$ in play. For example, if the teacher and student act in exactly the same manner regardless of the timestep: the teacher always sends $m_t = p_{ij}(o_t)$, and the student always makes the prediction $P(p_{ij}^{-1}(m_t)) = 1$.
This tells us that, in the case where two agents have trained with one another, the error metrics and performance metric are not sufficient to detect whether or not the teacher and/or the student are positively listening to the protocol being expressed within the episode, i.e. whether or not we can speak of \textit{intra-episodic protocol establishment}. To measure this we need to look at the agents responses to counterfactual protocols.
On the other hand, when interacting with a stranger, a high (mean) performance always indicates an intra-episodically established protocol.
When communicating with a stranger a high performance can only happen when, the TM error and SIC error are low and the protocol diversity is high. 
In this case, the teacher must be positively signalling $(m_t, o_t)$ and the student must be positively listening to messages $m_f$ and $m_t$.

\textbf{Protocol Responsiveness:} There are two senses in which an agent can be responsive to the protocol; \textit{student protocol responsiveness}, $R_S$, and \textit{teacher protocol responsiveness}, $R_T$. To measure these we expose the agent to random protocols and compute the following:
\begin{align}
    R_S = \exp\brac*{-\overline{\mathcal{L}_{SIC}}^*}, 
    ~\text{and}~ R_T = \exp\brac*{-\overline{\mathcal{L}_{TM}}^*}.
\end{align}
The additional notation over the error metrics indicates that we are measuring the mean of these values under the following special conditions. When measuring $R_S$, we put the agent in the role of the student and have it play with a synthetic teacher that generates and uses a random protocol. This isolates the student's responsiveness from the teacher's performance. For measuring $R_T$ we ignore the student all together and generate a random protocol by playing games where the mutation probability $p_m = 1$. 
When the teacher protocol responsiveness is high, 
then we know that the teacher is able to adapt to the protocol that was generated during the establishment phase.

We can speak of a protocol being established intra-episodically when the mutual information between the student's mental state and a previously received observation increases, after hearing a message the student has heard before. 
When protocols are generated at random and a final message is generated based on this protocol, 
then the student gets many different final messages. 
If the student has perfect responsiveness to these protocols, then we know that the student's mental state is affected by hearing a message that follows a protocol, and that the student is sensitively listening 
 to the final message.
Note, we here use the students prediction over the observation space as an indicator of this mutual information.


\textbf{Protocol Diversity}: This is a more interpretable variation on the protocol diversity error metric, and is only ever measured in an environment without message mutation, as message mutation can obscure seeing whether or not an agent has actually learned the ability to create an injective protocol. We compute this measure $P_D = \mathcal{L}_{PD}^{-1}$. The closer $P_D$ is to zero, the worse, the closer it is to one, the better.

\textbf{Zero-shot Cooperative Performance (ZCP):} For each of our different experimental set-ups we are interested in how agents from separate training environments are able to cooperate with one another. To assess this we concurrently train multiple agents, $A_1, \dots, A_k$, via self-play without any parameter sharing. We then take every combination of $A_i, A_j, j\neq i$ and run two sets of $170$ games, one set where $A_i$ is the teacher and $A_j$ is the student, and vice versa. The final zero-shot cooperative performance score is the mean performance across all of these games. Additionally, we refer to each combination of agents as a `stranger encounter'.

\textbf{Self-play Test Performance:} This is the mean performance when an agent plays with itself with the training parameters, e.g. if it played with a particular message mutation probability then it is tested with the same setting.
However, one difference is that the communication channel is no longer continuous.




\section{Experimental Results} \label{sec:experiments}




All experiments were written with Tensorflow 2 \citep{tensorflow2015-whitepaper} and can be found at: 
\url{https://www.github.com/DylanCope/zero-shot-comm}.
For all experiments, assume that $|\mathcal{O}_E| = 3$ and $|\Sigma| = 5$, unless otherwise stated. Each training epoch consists of 50 training steps.


\subsection{Baseline} \label{sec:baseline}

As a baseline we trained 6 agents with the loss function $\mathcal{L}_{AC}$. Each agent trained to minimal loss and achieved perfect self-play performance. After 30 stranger encounters we computed a zero-shot cooperative performance of $0.39 \pm 0.32$. This is close to the expected value from sampling answers from a uniform distribution, so we can conclude that these agents have not learned any capacity for zero-shot communication. Additionally, we did indeed find that a fixed protocol was established by each agent in their self-play environments.

\subsection{Protocol Fixing} \label{sec:protocol_fixing}

\begin{wrapfigure}{r}{0.45\textwidth}
    \centering
    \includegraphics[width=0.4\textwidth]{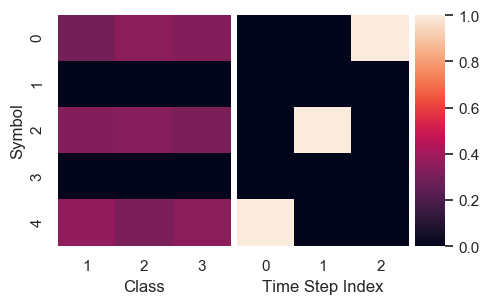}
    \caption{A visualisation of a temporally-fixed protocol}
    \label{fig:temporally_fixed_protocol}
\end{wrapfigure}

We observed that when training agents with the loss function $\mathcal{L}_{SIC} + \mathcal{L}_{TD} + \mathcal{L}_{PD}$ (and with no channel randomisation) there was convergence on fixed protocols, as would be expected. 
We also found that under certain conditions we could get the agents to consistently converge on \textit{temporally-fixed} protocols rather than \textit{observation-fixed} protocols. By this we mean that the teacher would always send the same ordered sequence of messages within each episode, regardless of the ordering of the observations, and they would adapt their final message appropriately. 



Figure \ref{fig:temporally_fixed_protocol} visualises a temporally-fixed protocol as a heat map where each cell shows the proportion of the time that a particular class or timestep index coincides with a particular symbol, e.g. we can see that symbol 4 can be sent alongside any class, but is only ever sent on the first timestep.
We noticed that a simple change to the agent's architecture can force temporally-fixed or observation-fixed protocols. 
In the Supplementary Material we analyse this phenomenon and speculate on why it occurs, however, for our purposes here we find that regardless of which strategies emerge in the absence of channel randomisation, when channel randomisation is applied there is convergence on protocols with observations as the subject.

\subsection{Effects of Message Mutation} \label{sec:msg_mut_experiments}

\begin{figure*}[!h]
\hspace{1.cm}
\begin{subfigure}{.45\textwidth}
    \centering
    \hspace{-1.75cm}
    \includegraphics[width=1\textwidth]{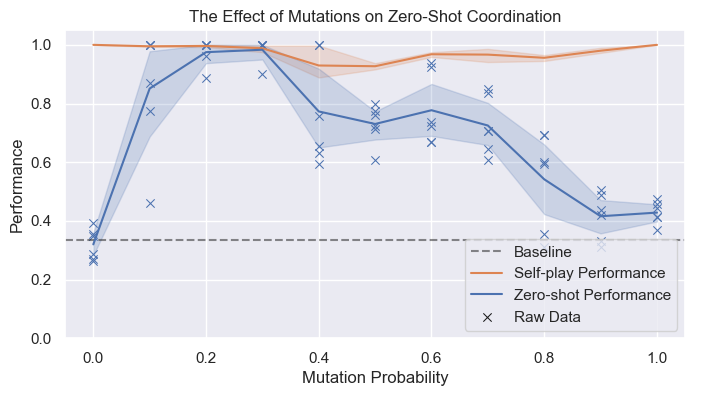}
    \caption{Results for message mutation experiment}
   \label{fig:msg_mut_1a}
\end{subfigure}
\begin{subfigure}{.45\textwidth}
  \centering
  \hspace{-1cm}
  \includegraphics[width=1\textwidth]{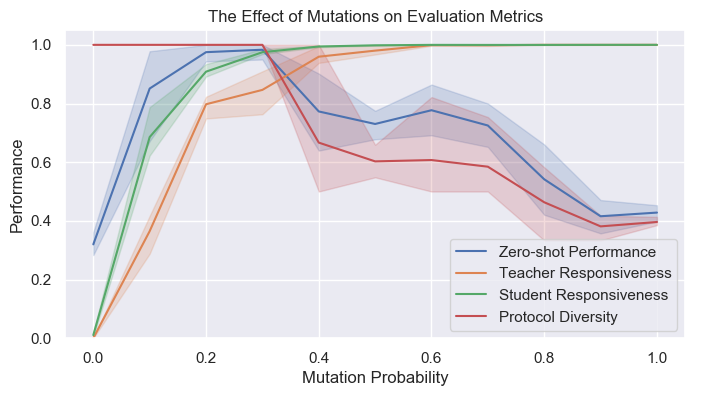}
    \caption{Evaluation metrics against mutation probability}
   \label{fig:msg_mut_1b}
\end{subfigure}%

\vspace{0.4cm}
\hspace{1.cm}
\begin{subfigure}{.45\textwidth}
    \centering
    \hspace{-1.75cm}
    \includegraphics[width=1\textwidth]{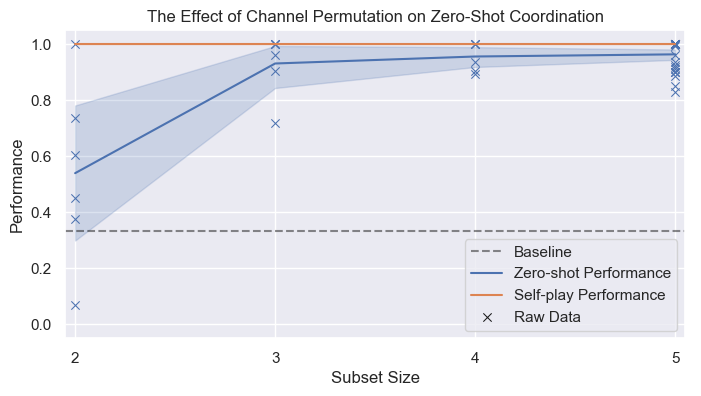}
\caption{Results for channel permutation experiment.}
   \label{fig:chan_perm_1a}
\end{subfigure}
\begin{subfigure}{.45\textwidth}
  \centering
  \hspace{-1cm}
  \includegraphics[width=1\textwidth]{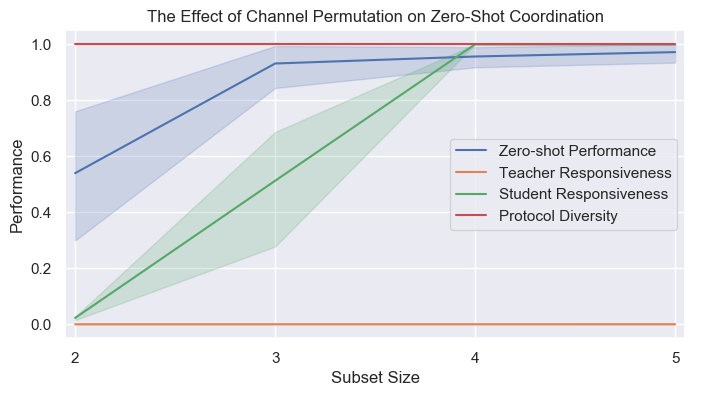}
\caption{Evaluation metrics against permutation subset size.}
  \label{fig:chan_perm_1b}
\end{subfigure}%
\caption{We visualise channel randomisation experiments.}
\label{fig:chan_perm_1}
\end{figure*}

In order to investigate the effects of message mutation we trained 3 agents for 11 different values of the mutation probability, resulting in 6 stranger encounters per value. 
As mentioned previously, message mutation requires acting on the protocol establishment phase, which in turn requires that the agents are incentivised to use the protocol being expressed in the first $|\mathcal{O}|_E$ timesteps. To do this we train the agents with the loss function $\mathcal{L}_{SIC} + \mathcal{L}_{TD} + \mathcal{L}_{PD}$.
In Figure \ref{fig:msg_mut_1a} we visualise the zero-shot cooperative performance (in blue). 
During training each agent takes both roles in the game with the mutation probability indicated on the $x$-axis,
but during the zero-shot evaluation the mutation probability is set to zero.
In order to make a fair comparison between the performance of agents trained with and without message mutation, we use the same evaluation environment without message mutation. 
In short, during self-play training there are 11 different levels of mutation probability,
but during the zero-shot evaluation there is no randomisation in the communication channel.

We can see a clear relationship between mutation probability and zero-shot performance, where the performance reaches a peak at $p_m = 0.3$ ($\text{ZCP} = 0.98 \pm 0.04$).
When the mutation probability is 0, so when there is no randomness involved, the zero-shot performance is poorest. This is as expected and verifies that the combination of loss functions is not enough on its own to encourage agents to learn the necessary skills for zero-shot communication, i.e. positive listening and signalling of the protocol itself.

Naively, one may assume that as the randomness in the environment increases, the performance increases, as the agents are forcefully exposed to more protocols. 
However, when the randomness is very large during training, the teacher may never learn to construct a protocol. This is reflected in the fact that when $p_m=1.0$, the protocol diversity measure $P_D$, is low (Table \ref{table:experiment_metrics}). Recall, that when measuring this metric we evaluate the agent in an environment without any channel randomisation, so it is unsurprising that the teacher has not learned skills for this domain. In other words, the teacher does not have to learn to create diverse protocols when mutation probability is 1, because the environment will always ensure an injective protocol.

We also look at the performance the agents had during self-play training, which is visualised in orange in Figure \ref{fig:msg_mut_1a}. We see that the performance during training is consistently high, and in particular for mutation probabilities 0 and 1. Lastly, we visualise a baseline, described in Section \ref{sec:baseline}. In Figure \ref{fig:msg_mut_1b} we see the zero-shot cooperative performance moving in concert with the protocol responsiveness measures and the protocol diversity measure. The $P_D$ starts high and remains high until around $p_m=0.3$, after which it starts to drop. The responsiveness starts low and monotonically increases with mutation probability. The ZCP peaks at the point where $P_D, R_S$, and $R_T$ are all high. This supports the argument that intra-episodic protocol establishment drives zero-shot communication.

\subsection{Effects of Channel Permutation}

We trained agents with channel permutation by using the $\mathcal{L}_{AC}$ loss function. In order to get the agents to reliably converge we found that we needed to use a temperature annealing schedule on the communication channel \citep{Jang2017CategoricalGumbel-Softmax}. We used an exponential decay schedule where the temperature starts at 10, updates once an epoch, and ends at 0.1 at epoch 200. After which it stays constant at this value.
After training 6 agents with permutation over all symbols, we ran 30 stranger-encounters and found a mean zero-shot test performance of  $0.96 \pm 0.05$. 

In Figure \ref{fig:chan_perm_1a} we visualise the zero-shot cooperative performance (in blue) and self-play performance (in orange) across various subset sizes. We do not show results for subset size zero or one as these are functionally equivalent to no channel randomisation, and are thus represented by the baseline. On the left, at subset size 2, we find that there is some improvement, but a very high variance. As we move to the right we see that the variance decreases and the performance approaches perfect play. But this does not come without any cost; we find that as the subset size grows so does the number of training steps needed for the system to converge. Fortunately, this growth is not too dramatic; it takes roughly 130 epochs to converge with $k=4$ and 150 epochs for $k=5$. See the Supplementary Material for full training histories of each of the experiments. In Figure \ref{fig:chan_perm_1b} we see that student responsiveness goes up as the amount of permutation increases, and that the protocol diversity is high for any level of permutation.  
However, we also see that teacher responsiveness constantly remains low, which means  
that the teacher is not able to, given a random protocol,  send a corresponding correct final message.
In this environment the teacher can get a high performance even if it always uses the same protocol.



\subsection{Comparing Experiments}

As we can see from comparing \ref{fig:msg_mut_1a} and \ref{fig:chan_perm_1a}, both channel randomisation methods can dramatically improve the zero-shot cooperative performance. While both methods can be tuned by a hyperparameter -- mutation probability vs. subset size -- we see in the case of message mutation that there is a balancing act that needs to be performed between exposure to new protocols and inhibiting the teacher from learning how to develop good protocols. Our finding that the optimum mutation probability is around 0.3 cannot be assumed to hold in other environments. On the other hand, increasing the channel permutation subset size does not introduce any similar such trade-off. It also has the advantage that it does not require any specific training signals that explicitly reference the adherence to the protocol expressed within the episode, i.e. it does not require manual identification of the protocol establishment phase. 

\begin{wraptable}{r}{0.6\textwidth}
\centering
\begin{tabular}{@{}lllll@{}}
\toprule
                     & $R_T$ & $R_S$ & $P_D$ & ZCP               \\ \midrule
Baseline             & 0    & 0    & 1    & $0.39$ \\
Permutation ($k=5$)  & 0    & 1    & 1    & $0.96$ \\
Mutation ($p_m=0.3$) & 0.85 & 0.97 & 1    & $0.98$ \\ 
Mutation ($p_m=1.0$) & 1    & 1    & 0.38 & $0.49$ \\ \bottomrule
\end{tabular}

\vspace{0.2cm}
\caption{Mean metrics from different experiments} \label{table:experiment_metrics}
\end{wraptable}

However, channel permutation is not without its disadvantages. Firstly, we observed that training times were significantly longer for channel permutation, and there was more tweaking necessary to get the system to reliably converge. But this should not be taken as the final word that no training regime for channel permutation could be more efficient, as we did not systematically explore different hyperparameters (learning rate, temperature annealing schedules, etc.). Additionally, we should be apprehensive about comparing training times between the message mutation and channel permutation experiments because they were trained with different loss functions. It is reasonable to expect the message mutation systems to converge faster due to the loss functions providing much more training signal. Secondly, and more importantly, channel permutation does not lead to teachers that pay attention to the protocol that they are communicating, although this skill is not strictly necessary for zero-shot communication. This is shown in Table \ref{table:experiment_metrics} where we see that the teacher protocol responsiveness, $R_T$, is zero, compared to 0.85 for message mutation ($p_m=0.3$). 




\section{Discussion} \label{sec:discussion}

\subsection{Summary}

The central hypothesis in this work is that channel randomisation methods have an effect on learning the skill of intra-episodic protocol establishment, which can be necessary for zero-shot communication. We demonstrated this effect empirically by showing that our proposed methods of message mutation and channel permutation indeed have the expected outcomes in our test environment.

We introduced a series of new definitions and measures to aid our investigation. Firstly, the definitions of intra- vs. inter-episodic protocol establishment were given formally in terms of causal influence (Section \ref{sec:intra-episodic}). Secondly, our error and evaluation metrics (Section \ref{sec:measures}) provided concrete ways of detecting the presence of these phenomena. Precisely, the responsiveness measures evaluate how agents react to counterfactual protocols expressed within an episode, and the protocol diversity metric assesses whether the protocol is injective or not. Only in conjunction do these metrics identify the ability to intra-episodically establish a protocol, when analysing an agent on its own. 
When analysing interactions between strangers, high performance alone is enough to identify this ability (in our set-up).
We can measure the presence of fixed protocols by poor zero-shot performance and high self-play performance, or by low protocol responsiveness and high self-play performance.

Figure \ref{fig:msg_mut_1a}, where $p_m=0$, and the baseline experiment (Section \ref{sec:baseline}), show that the absence of randomisation led to poor zero-shot communication performance. This was accompanied by the convergence on different inter-episodically established protocols by each of the independently trained agents.  As channel permutation was more strongly applied (Figure \ref{fig:chan_perm_1a}) we see that zero-shot performance increases in concert. 
On the other hand, more message mutation only helps up until it interferes with the agent's opportunity to learn to produce good protocols. Figure \ref{fig:msg_mut_1b} clearly shows that as mutation probability increases, so does responsiveness to the protocol, but after a threshold the protocol diversity measure starts to decrease, and with it the zero-shot performance.








\subsection{Generalizability of our Work}

We expect similar zero-shot performance may be achieved through training multiple agents in parallel and matching agents in some sort of league.
An advantage of our approach is that our method of randomising only requires us to train two networks, rather than a league of networks.
The number of bits needed to store all protocols possible with $| \Sigma |= N$ symbols and  $|X| =k$ subjects grows exponentially, with $(N-k)^{k}$ as lower bound,
whereas the number of bits available in a network grows linearly in the number of weights, see Appendix.
This makes it unfeasible to store all possible protocols in the weights for large $N$ and $k$, and agents performing well during training will have to have learned the general skill of adapting to a new protocol.
We think that, in general, some form of protocol needs to be established for zero-shot communication, and that randomisation in the communication channel during self-play can help with generalisation to other players.
However, it may be difficult to apply our proposals on an arbitrary communication channel.

\subsection{Further Work}

In order to assess the scalability of our proposals, they could be transported to more complex domains, such as Sequential Social Dilemmas \citep{Leibo2017Multi-AgentDilemmas}. For channel permutation, this should be a relatively straightforward implementation. However, for message mutation it is necessary that we can identify where in the episode the protocol ought to be established, i.e. the protocol establishment phase.
Furthermore, this identification necessity applies to all non-performance based metrics.


There are several aspects of the set-up presented in this work that could be subjected to further empirical scrutiny. For example, we used a relatively small number of possible environment observations ($|\mathcal{O}_E| = 3$) and a relatively small number of possible messages ($|\Sigma| = 5$). This choice was somewhat arbitrary and was made to reduce training time. As a result our message mutation proposal required using `kind mutations' that did not threaten to unfairly damage protocol diversity. This was an additional implementation burden that we expect would not need to be taken on if $|\Sigma| \gg |\mathcal{O}_E|$. 
Another approach that we did not investigate is whether or not combining our proposals has any interesting effects, or 
whether or not sometimes turning off message mutation would encourage the teacher to learn protocol generation skills for high values of $p_m$.


\bibliographystyle{ACM-Reference-Format} 
\bibliography{refs}


\clearpage
\appendix

\section{Appendix}

\subsection{Alternative positive listening definitions}

We can dissect positive listening based on which type of action is changed by the listening.
\begin{definition}[Decisive, Communicative, and Private Listening]
Consider an agent $i$ with action space $\mathcal{A}_i = \mathcal{A}_i^e \times \Sigma \times \mathcal{Z}_i$ and policy $\pi_i: \mathcal{O}_i \times \Sigma \rightarrow \mathcal{A}_i$. We decompose the policy into three different policies and use the intuitive notation 
$\pi_{ie}: \mathcal{O}_i \times \Sigma \rightarrow \mathcal{A}_i^e$,  
$\pi_{i\sigma}: \mathcal{O}_i \times \Sigma \rightarrow \Sigma$, and
$\pi_{iz}: \mathcal{O}_i \times \Sigma \rightarrow \mathcal{Z}_i$.
We say that agent $i$ is respectively decisive, communicative or private listening to another agent $j$ at time $t$ if $j$ has just sent message $m_{ijt} \in \Sigma$ to $i$ and $\pi_{ie}(o_{it}, m_{ijt}) \neq \pi_{ie}(o_{it}, m_{\mathnull})$, 
$\pi_{i\sigma}(o_{it}, m_{ijt}) \neq \pi_{i\sigma}(o_{it}, m_{\mathnull})$ or 
$\pi_{iz}(o_{it}, m_{ijt}) \neq \pi_{iz}(o_{it}, m_{\mathnull})$.
\end{definition}

\subsection{Additional Justification for Our Environment}

In our design discussions, we toyed with adding more complexity to our environment, however, we settled on the task in the paper for a few reasons. Key ways in which the task could be more complex are: (1) a task where communication is instrumental, but not necessary, and (2) more complex tasks where communication is essential. For (1), as discussed in our introduction, we are interested in tasks where communication is instrumental. However, in our environment communication is the end-goal, because otherwise, we would unnecessarily obscure measurement of communication. As \cite{Lowe2019OnCommunication} point out, such measurement can be very tricky, even in simple environments. Our environment allowed us to easily measure positive signalling and positive listening, not only for the observations but for the protocol itself (see $R_S$ and $R_T$). 

For (2): our environment demonstrates a scenario in which agents learn zero-shot generalisable communication strategies. This does not necessarily have implications for all environments in which communication is instrumental or necessary. Indeed, our environment assumes that there is opportunity for a protocol establishment phase. However, we argue, that our particular classification task could have been replaced with more complex tasks (where communication is necessary). For instance, suppose the student solves a maze blind-folded, while the teacher sees the maze. Despite this new environment being more interesting, we do not see how adding such complexity would further justify our claims about communication. We would just be adding an image classification task and control task on top of our key communication task. DNNs are already well-documented to deal with these problems.

\subsection{Formal Construction of the Environment}

Given a student $S$, a teacher $T$, and a communication channel $\sigma: \mathcal{A}_T \rightarrow \Sigma$, we can formally define our environment as a Markov game:
\begin{align}
    &\mathcal{S} = \curlybrac*{
        (t, o_E, m_T, H) \in \mathcal{S}^\infty
        ~:~ 0 \leq t \leq |\mathcal{O}_E|
    } \\ 
    &\mathcal{S}^\infty = 
        \mathds{N} \times \mathcal{O}_E \times (\Sigma \cup \{m_\mathnull\}) \times \mathcal{O}_E^*
    \\
    &\mathcal{S}_T = \curlybrac*{
        (|\mathcal{O}_E|, o_E, m_T, H) \in \mathcal{S}
    } \\
    &\rho = \text{Uniform}(\curlybrac*{
        (0, o_E, m_\mathnull, \mathnull) ~:~ o_E \in \mathcal{O}_E
    }) \\
    &\mathcal{A}_S = \Delta(\mathcal{O}_E),~ \mathcal{O}_S = \mathcal{O}_E \times \Sigma\\
    &\mathcal{A}_T = \mathds{R}^{|\Sigma|},~ \mathcal{O}_T = \mathcal{O}_E \times \Sigma \\
    &\mathcal{T}((t, o_E, m_T, H), y_S, m_T) = \text{Uniform}(\mathcal{S}_{next}) \\
    &\mathcal{S}_{next} = \curlybrac*{
        (t+1, o_E', \sigma(m_T), Ho_E) :  o_E' \in O
    } \\
    &O = \begin{cases}
        \mathcal{O}_E / (Ho_E) & \text{if}~ t < |\mathcal{O}_E| - 1 \\
        H & \text{if}~ t = |\mathcal{O}_E| - 1 \\
        \{o_\mathnull\} & \text{otherwise} \\
    \end{cases} \\
    &\omega_S((t, o_E, m_T, H)) = \begin{cases} 
        (o_E, m_T) & \text{if}~ t < |\mathcal{O}_E| - 1 \\
        (o_\mathnull, m_T) & \text{otherwise}
    \end{cases} \\
    &\omega_T((t, o_E, m_T, H)) = (o_E, m_T)
\end{align}



\subsection{Visualisations of episodes}

\begin{figure*}
\begin{subfigure}{.45\textwidth}
  \centering
    \includegraphics[width=0.9\textwidth]{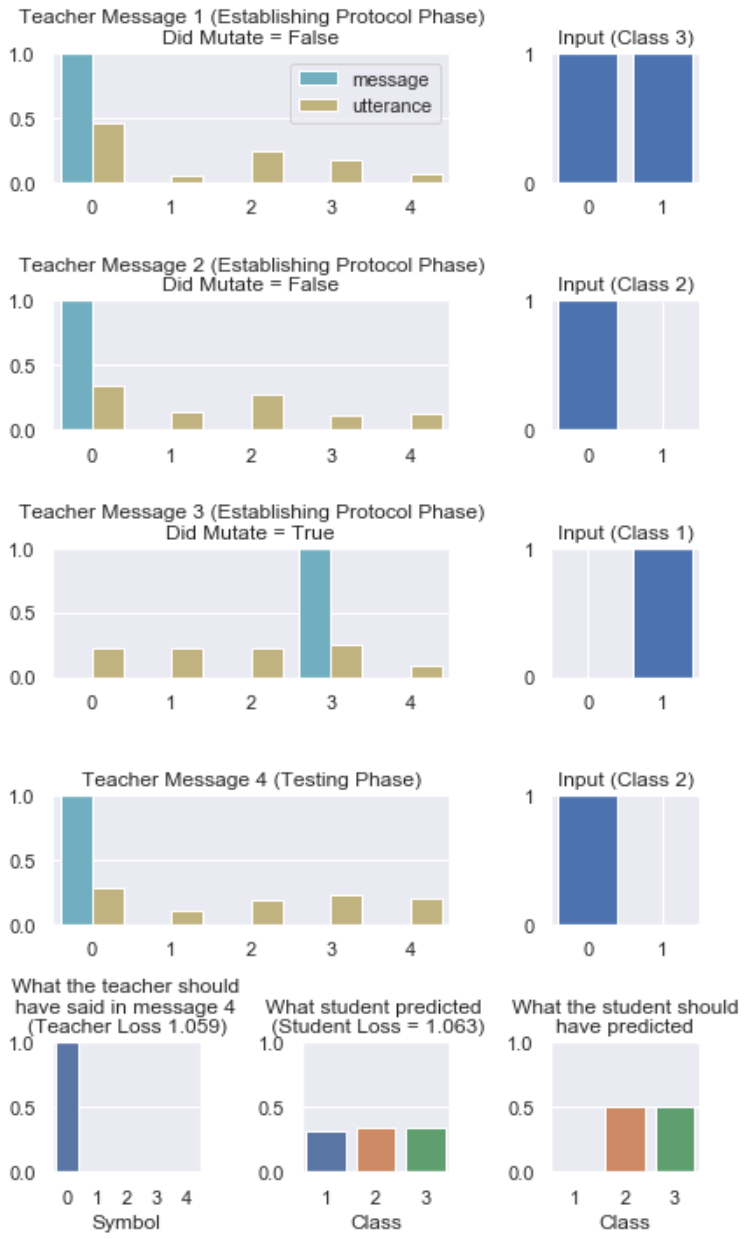}
  \caption{Poor performance from untrained agents}
  \label{fig:game_examples_a}
\end{subfigure}
\begin{subfigure}{.45\textwidth}
  \centering
    \includegraphics[width=0.9\textwidth]{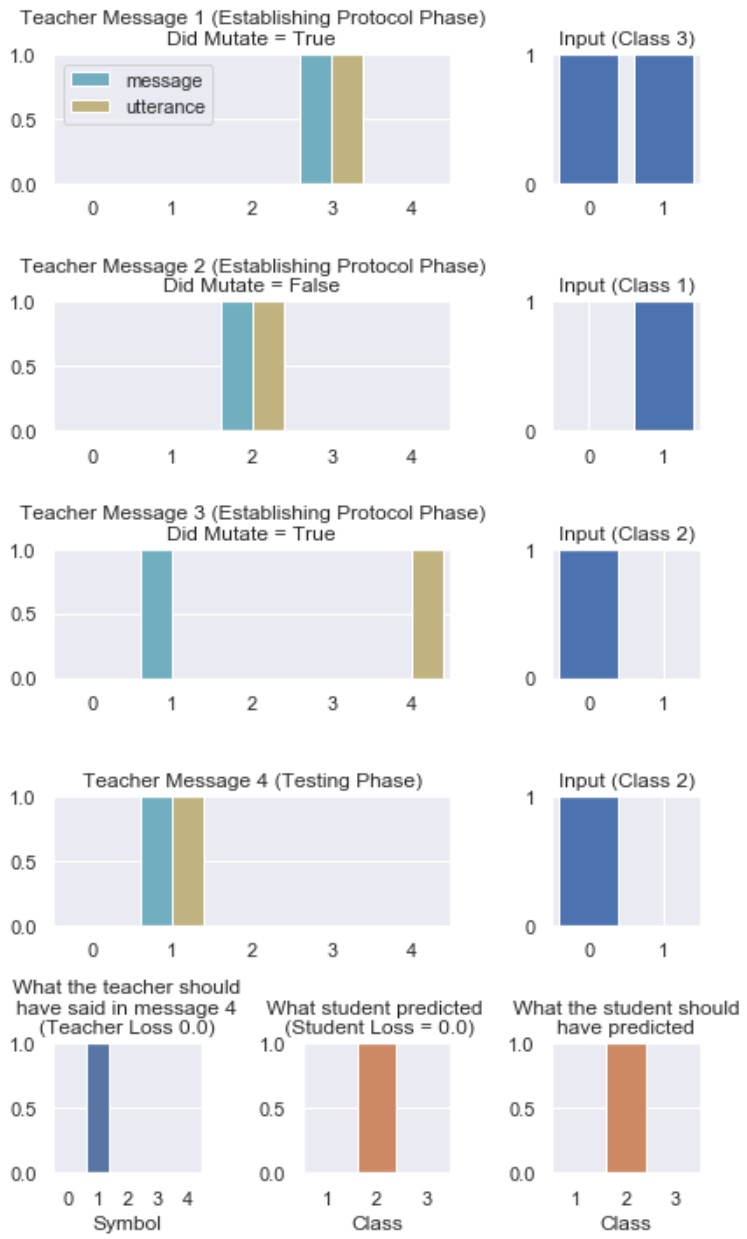}
  \caption{Perfect performance from trained agents}
  \label{fig:game_examples_b}
\end{subfigure}%
\caption{Visualisations of episodes in our environment. The first four rows of each subfigure represent the four timesteps of the episode. The final row demonstrates the outcomes in terms of what the agents should have done and what predictions the student made. In the left-hand column of each timestep we can see the teacher's utterance in yellow and what was sent as a message in teal. In the right-hand column of each timestep we can see the environment observations and their class labels. Keep in mind that the first three observations are made by both the teacher and the student, but the last is only observed by the teacher.}
\label{fig:game_examples}
\end{figure*}

See Figure \ref{fig:game_examples}.

\subsection{Training Details}

For each of our experiments we use an RMSprop optimiser with a learning rate of 0.01 and a decay factor of 0.9 \citep{Hinton2012NeuralDescent, tensorflow2015-whitepaper}. We use a batch size of 32.

Each agent $i$'s policy RNN is composed of a dense layer with 128 hidden units, an LSTM layer with 64 hidden units, and a final dense layer with $|\mathcal{O}_E|+|\Sigma|$ units. The final layer had no activation function, the LSTM cell had a tanh feed-forward activation and a sigmoid recurrent activation, and the activation on the first layer is discussed in Section \ref{sec:protocol_fixing}. The input to the first layer is the concatenation of the three aforementioned inputs. The class predictions are extracted from the final layer by slicing out the first $|\mathcal{O}_E|$ units and applying a softmax function, and the utterance is extracted by slicing out the remaining $|\Sigma|$ units.

\begin{figure*}[!b]
\hspace{1.cm}
\begin{subfigure}{.45\textwidth}
    \centering
    \hspace{-1.75cm}
    \includegraphics[width=1\textwidth]{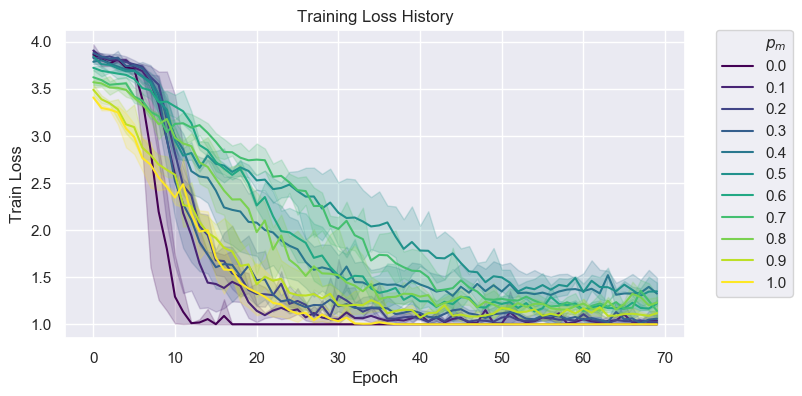}
    \caption{Training history for message mutation experiments}
   \label{fig:msg_mut_train_history}
\end{subfigure}
\begin{subfigure}{.45\textwidth}
  \centering
  \hspace{-1cm}
  \includegraphics[width=1\textwidth]{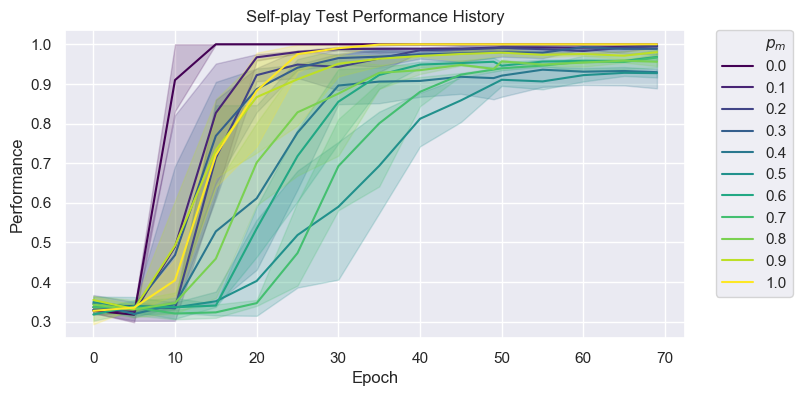}
    \caption{Test performance history for message mutation experiments}
   \label{fig:msg_mut_test_history}
\end{subfigure}%

\vspace{0.4cm}
\hspace{1.cm}
\begin{subfigure}{.45\textwidth}
    \centering
    \hspace{-1.75cm}
    \includegraphics[width=1\textwidth]{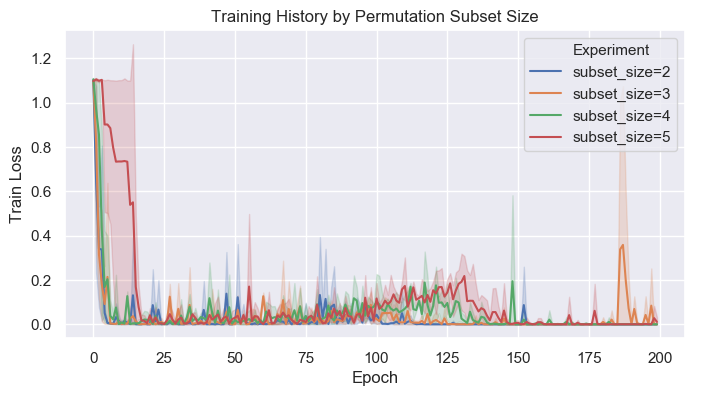}
\caption{Training history for channel permutation experiments}
   \label{fig:chan_perm_train_history}
\end{subfigure}
\begin{subfigure}{.45\textwidth}
  \centering
  \hspace{-1cm}
  \includegraphics[width=1\textwidth]{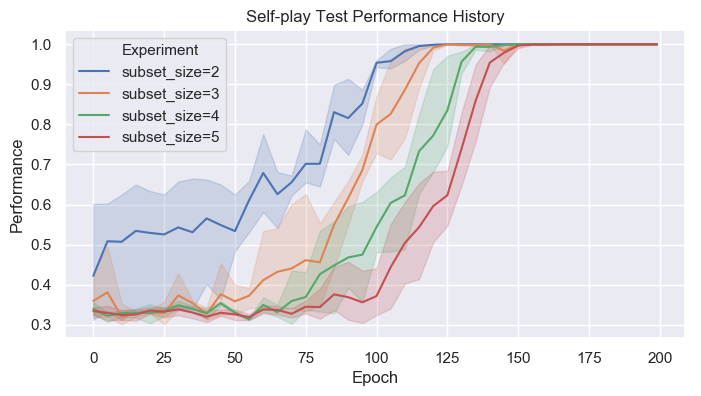}
\caption{Test performance history for message mutation experiments}
  \label{fig:chan_perm_test_history}
\end{subfigure}%
\caption{Training histories}
\label{fig:training_histories}
\end{figure*}

Figure \ref{fig:training_histories} shows the training histories for each of the different experiments.

\subsection{Protocol Fixing}

Following from the observation made in Section \ref{sec:protocol_fixing}, in both cases the agents' policy networks were composed of three layers; a dense layer, an LSTM cell, and another dense layer. For the temporally-fixed (TF) agents, the first layer had a ReLU activation, and for the observation-fixed (OF) agents, it had no activation function.




We show that, if there is a linear activation function, then the agent can only learn an OF strategy.
Whereas, if there is a non-linear activation function, then both the TF and OF strategies are possible.

To understand why the architecture leads agents to choose one fixing paradigm over another, consider the computational steps that the teacher must take in order to be a TF-agent. 
It must, at each timestep discriminate between different categories of observations, and encode this information in its hidden state.
As elements of $\mathcal{O}_E$ are binary vectors no single-layer network can discriminate between all of the classes \citep{minsky2017perceptrons}. Without a non-linear activation function before the LSTM cell there is effectively only one linear transformation and one activation function before the hidden state. As such, it is impossible for the agent to encode the observation classification into the hidden state.




This is of interest because it shows that given a non-linear activation function, the incentives can be enough on their own to lead to the emergence of strategies that involve some small degree of responsiveness to the protocol expressed within the episode on behalf the student ($R_S = 0.05$ for the TF-Agent), despite there existing strategies that do not require this skill. However, they are not enough to lead to good zero-shot performance, as both the OF and TF agents performed around the baseline when paired with strangers.


\subsection{Theoretical Analysis}

\subsection*{Protocol-fixing under channel permutation}


Here we show that it may be possible for agents to invert the permutation map rather than learn strategies that are independent of the map.

\begin{enumerate}
    \item Suppose that a teacher and a student have a predefined protocol over a class of observations, $p_\mathcal{O}: \mathcal{O} \rightarrow \Sigma$, that they wish to use for communication.
    \item However, there is an unknown random permutation map, $f: \Sigma \rightarrow \Sigma$, in the communication channel
    \item So the agents agree beforehand for the teacher to send all possible symbols to the student in a fixed order, taking $|\Sigma|$ timesteps. In other words, the agents have memorised a second protocol over timesteps, $p_T : \mathds{N} \rightarrow \Sigma$. 
    \item For each timestep $0 \leq t < |\Sigma|$ the teacher sends $p_T(t)$ and the student receives and memorises $f(p_T(t))$. At $t = |\Sigma|$, the student knows the value of $f(s) ~\forall s \in \Sigma$.
    \item So when the teacher makes observation $x \in \mathcal{O}$ they can send message $p_\mathcal{O}(x)$ with the knowledge that the student can compute $x = p_\mathcal{O}^{-1}(f^{-1}(s))$. Given that both $f$ and $p_\mathcal{O}$ are bijective, we know that their inverses can be found.
\end{enumerate}

\subsection*{Feasibility of Memorising All Protocols}

When agents inter-episodically established a protocol, this converged protocol must be somehow encoded in the weights of the policy network. 

Through randomisation, we can expose agent to many protocols during training, and 
we expect this will incentivise them to develop the general skill of interpreting any new protocol. Alternatively they could store all (or many) possible protocols, and use the protocol establishment phase to communicate which protocol they will be using during this session. 
To prevent this we could make the number of possible protocols so large, that it becomes infeasible for an agent to memorise all possible protocols.
We do a back-of-the-envelope calculation to assess the feasibility of storing all possible protocols.

\textbf{Number of Possible Protocols}: 
The first question we ask ourselves is, given a set of symbols $\Sigma$ with cardinality $| \Sigma |= N$ and a set of subjects $X$ with cardinality $ |X| = k$, how many possible protocols are there?
We assume that each protocol is an injective map from $X$ to $\Sigma$,
in which case there are $\frac{N!}{(N-k)!}$ possible protocols.




\textbf{Number of bits needed to encode all protocols}: 
Given $N$ symbols and $k$ objects we can choose the following encoding for an injective function $f \colon X \rightarrow \Sigma$. We can encode each element of $\Sigma$ using $\log_2(N)$ bits. We assume an ordering on $X$ and we associate an image to each element of $X$ using $\log_2(N)$ bits. In total each injective function needs $k \cdot \log_2(N)$ bits to be encoded using the above encoding.

Encoding $\frac{N!}{(N-k)!}$ many functions then takes $\frac{N!}{(N-k)!} \cdot k \cdot \log_2(N)$ bits. 
Note that $(N-k+1)^k < \frac{N!}{(N-k)!} < N^k$. Moreover, $(N-k)^k < \frac{N!}{(N-k)!} \cdot k \cdot \log_2(N) < N^{k+1}$. We conclude that $(N-k)^k$ is a lower bound for the number of bits we need to encode all possible protocols (using the encoding that we laid out).

For the numbers $N=5$ and $k=3$ that we used in our experiments this lower bound is not very large, but for example, suppose that $N= 20$ and $k=10$, then the number of bits needed to encode all possible protocols, becomes 10 billion.

\textbf{Feasibility of storing protocols}: 
One weight in a neural network is 32 bits. 
A network with $W$ weights and biases contains $W*32$ bits.
If $W*32$ is smaller than $(N-k)^k$, then it’s not possible to express all the functions in the network

In our set-up, at most one protocol is introduced per episode.
If the number of training episodes is smaller than the number of possible protocols, 
then it is unfeasible to store all protocols in the weights during training.

\end{document}